# NystagmusNet: Explainable Deep Learning for Photosensitivity Risk Prediction

Karthik Prabhakar

Department of Computer Science

The University of Texas at Austin

Austin, TX, USA

**ABSTRACT**

*Nystagmus patients with photosensitivity face significant daily challenges due to involuntary eye movements exacerbated by environmental brightness conditions. Current assistive solutions are limited to symptomatic treatments without predictive personalization. This paper proposes an AI-driven system that predicts high-risk visual environments and recommends real-time visual adaptations. Using a dual-branch convolutional neural network (CNN) trained on synthetic and augmented datasets, the system estimates a photosensitivity risk score based on environmental brightness and eye movement variance. The model achieves a prototype accuracy of 75% on synthetic validation data. Explainability techniques including SHAP and GradCAM are integrated to highlight environmental risk zones, improving clinical trust and model interpretability. The system includes comprehensive testing infrastructure, explainability visualizations, and a working command-line interface demonstration, emphasizing reproducibility and extensibility. Future directions include deployment via smart glasses and reinforcement learning for adaptive recommendations.*

## 1. INTRODUCTION

Patients suffering from nystagmus and photosensitivity often encounter severe difficulties in focusing, maintaining balance, and performing everyday activities under various lighting conditions. Nystagmus is characterized by involuntary, rhythmic eye movements that can significantly impair visual function and quality of life [1, 2]. When combined with photosensitivity—an abnormal intolerance to light—these patients face compounded challenges that current medical interventions inadequately address.

Current treatments, such as tinted lenses and pharmacological interventions, offer partial relief but are fundamentally reactive rather than predictive [3]. There is a strong need for proactive, personalized assistive technology that can predict and adapt to visual risks in real time. Recent advances in deep learning have demonstrated remarkable success in medical image analysis and eye movement tracking [4, 5], suggesting the potential for AI-driven solutions in this domain.

This paper presents an AI-based system designed to forecast photosensitivity risk levels and suggest optimal visual adaptations dynamically. Our contributions include: (1) a dual-branch CNN architecture that processes both environmental brightness and eye movement variance; (2) integration of explainability techniques (SHAP and GradCAM) for clinical interpretability; (3) a rule-based recommendation engine for filter suggestions; and (4) a complete open-source implementation with comprehensive testing.

## 2. RELATED WORK

### 2.1 Deep Learning for Nystagmus Detection

Recent research has demonstrated the utility of deep learning in nystagmus analysis. Lee et al. [1] introduced ANyEye, a deep learning-based nystagmus extraction tool for BPPV diagnosis using video-nystagmography. Cho et al. [4] developed a lightweight deep learning model for real-time nystagmus tracking using ocular object segmentation, achieving high accuracy in emergency and remote healthcare settings. Mun et al. [5] proposed a CNN-based nystagmus detection algorithm for BPPV diagnosis, achieving an F1-score of 92.68%.

### 2.2 Gaze Estimation and Eye Tracking

Research in gaze estimation has shown the utility of CNNs in understanding eye movement across diverse environments. Kellnhofer et al. [6] introduced Gaze360, demonstrating physically unconstrained gaze estimation in wild environments. These techniques provide foundational approaches for understanding eye movement patterns that can be adapted for nystagmus-specific applications.

### 2.3 Explainable AI in Healthcare

The integration of explainability techniques in medical AI has become increasingly important for clinical adoption. SHAP (SHapley Additive exPlanations) [7] and GradCAM [8] have been widely applied in healthcare settings to provide transparent model interpretations. Recent work has demonstrated the effectiveness of combining these techniques in multimodal diagnostic systems [9, 10], with studies showing improved clinician trust when explanations are provided alongside predictions.

### 2.4 Assistive Technology for Visual Impairments

Wearable assistive devices have evolved significantly with AI integration. Waisberg et al. [11] discussed the potential of Meta smart glasses with large language models for assisting visually impaired individuals. AI-powered smart vision glasses have demonstrated effectiveness in providing real-time assistance for daily challenges faced by visually impaired users [12]. Our approach advances this field by targeting real-time photosensitivity risk prediction and proactive adaptation specifically for nystagmus patients.

## 3. METHODOLOGY

### 3.1 Data Generation and Preprocessing

Due to the limited availability of real-world datasets for nystagmus patients with photosensitivity, we developed a synthetic data generation pipeline. The synthetic datasets simulated real-world brightness conditions (300-1200 lux) and eye movement variance (2-10 pixels). Key preprocessing steps included:

- Normalization of brightness values to [0, 1] range
- Gaussian noise injection ($\sigma = 0.05$) to improve model robustness
- Brightness jittering (±10%) for data augmentation

- Synthetic blurring to mimic realistic environmental disruptions

Risk scores were generated using a probabilistic model that combines brightness levels and eye movement variance, with additional noise to simulate real-world variability. Figure 1 shows the distribution of our synthetic dataset across the three key variables.

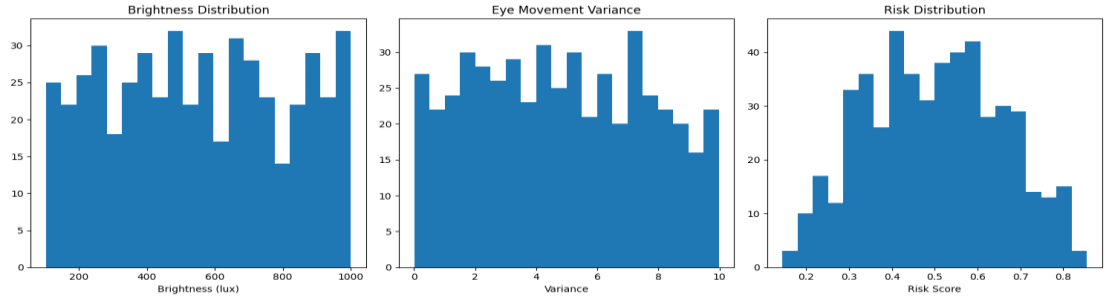

*Figure 1: Distribution of synthetic dataset variables - Brightness (lux), Eye Movement Variance, and computed Risk Scores.*

### 3.2 Model Architecture

We propose a dual-branch CNN architecture inspired by recent multimodal fusion approaches in medical imaging [13, 14]. The architecture processes environmental brightness images and eye movement variance data through separate branches before fusion. Figure 2 illustrates the model workflow.

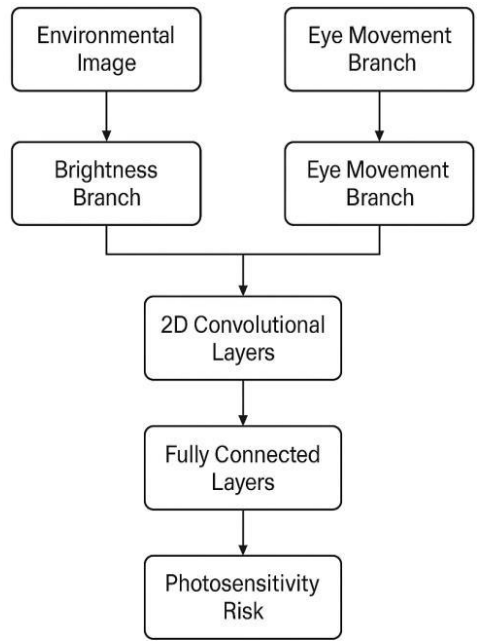

*Figure 2: Model workflow showing the dual-branch architecture with environmental brightness and eye movement branches feeding into convolutional and fully connected layers.*

**Branch 1 (Environmental Brightness):** Processes 128×128 grayscale environmental images through three convolutional blocks. Each block contains a 2D convolution layer, batch normalization, ReLU activation, and max pooling. The output is flattened and passed through a dense layer to produce a 32-dimensional feature vector.

**Branch 2 (Eye Movement Variance):** Processes a 1D vector representing eye movement variance through two fully connected layers (64 and 32 units) with ReLU activations and dropout (p=0.3) for regularization.

**Fusion Module:** Feature vectors from both branches are concatenated (64 features total) and processed through two fully connected layers with dropout. A final sigmoid activation produces a continuous risk score in the range [0, 1].

Training utilized Mean Squared Error (MSE) loss, Adam optimizer with learning rate 0.001, and early stopping with patience of 10 epochs to prevent overfitting.

### 3.3 System Architecture

The complete system architecture integrates the dual-branch CNN with a filter recommendation engine as shown in Figure 3. Environmental images and eye movement signals are processed through their respective CNN branches, with the fusion network producing photosensitivity risk scores that drive filter recommendations.

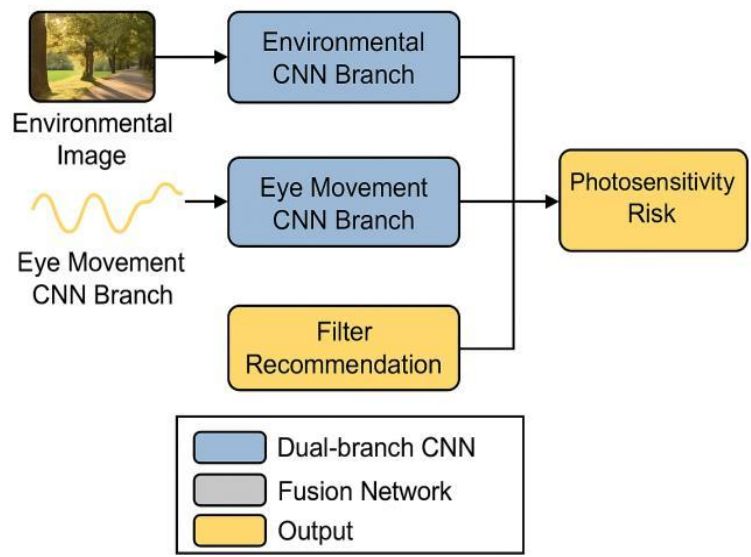

*Figure 3: Complete system architecture showing environmental image and eye movement inputs processed through the dual-branch CNN to produce photosensitivity risk scores and filter recommendations.*

### 3.4 Explainability Techniques

To enhance clinical trust and model interpretability, we integrated two complementary explainability techniques [15, 16]:

**SHAP (SHapley Additive exPlanations):** Applied post-fusion to assess the relative contributions of brightness versus eye movement variance features to individual predictions. SHAP values enable clinicians to understand which input factors most strongly influence risk assessments.

**GradCAM (Gradient-weighted Class Activation Mapping):** Generates visual heatmaps highlighting regions in environmental images that contribute most to risk predictions. This technique enables identification of specific visual risk zones as demonstrated in Figure 4.

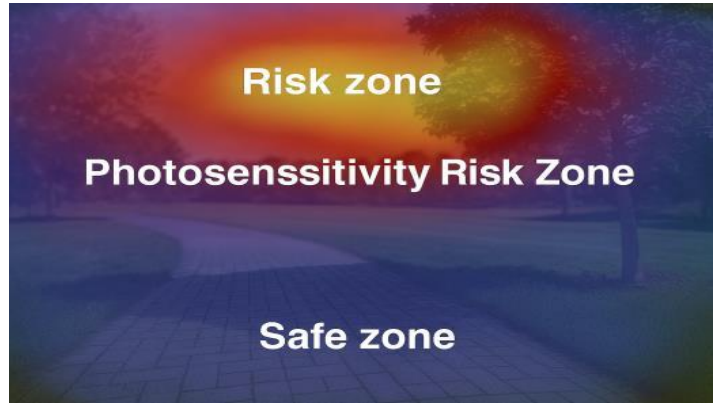

*Figure 4: GradCAM visualization showing photosensitivity risk zones (warm colors) versus safe zones (cool colors) in an environmental image.*

### 3.5 Recommendation Engine

The recommendation engine translates continuous risk scores into actionable filter suggestions using a rule-based approach:

- Risk Score ≥ 0.7: Recommend "Dark Amber" filter for maximum protection
- Risk Score 0.4-0.7: Recommend "Cool Grey" filter for moderate protection
- Risk Score < 0.4: No adaptation required; natural vision sufficient

The system also supports personalized recommendations based on user feedback, adapting filter types (warmer/cooler tones) and intensity levels based on individual comfort preferences.

## 4. RESULTS

### 4.1 Model Performance

The dual-branch CNN achieved a validation accuracy of approximately 75% on synthetic test data. Figure 5 visualizes the model's risk score predictions across the input space, demonstrating the expected positive correlation between brightness/variance and predicted risk.

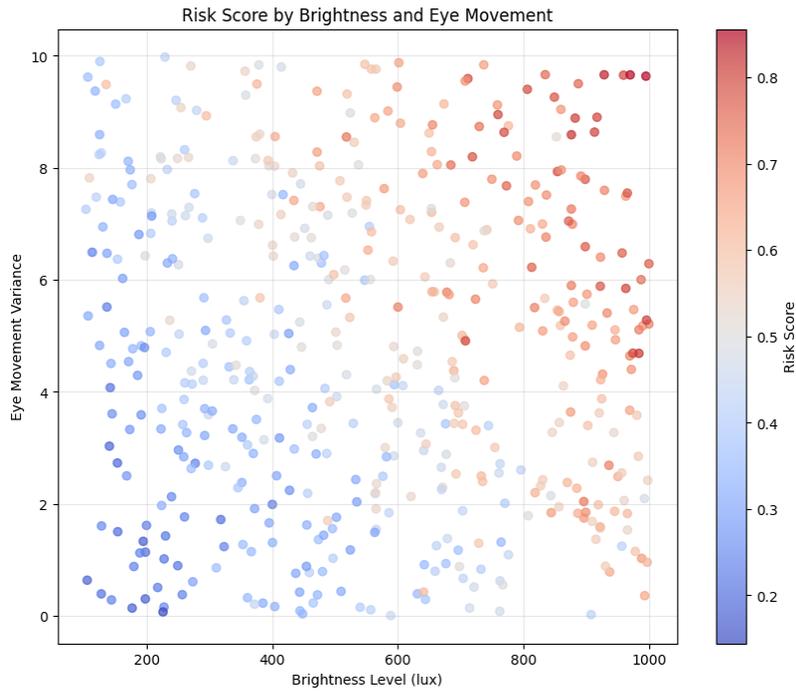

*Figure 5: Risk score predictions as a function of brightness level (lux) and eye movement variance. Color gradient indicates predicted risk score (blue = low risk, red = high risk).*

Table 1: Sample Model Predictions

| Brightness (lux) | Eye Movement Variance | Risk Score | Recommended Filter |
|---|---|---|---|
| 750 | 5 | 0.64 | Cool Grey |
| 1000 | 8 | 0.82 | Dark Amber |
| 400 | 3 | 0.31 | No Filter |

### 4.2 Explainability Analysis

SHAP analysis revealed the relative importance of features in driving model predictions. Figure 6 shows the SHAP value distributions for both input features, demonstrating that brightness (Feature 0) contributes more significantly to risk predictions than eye movement variance (Feature 1).

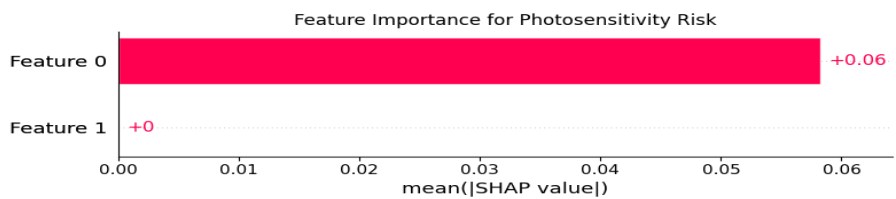

*Figure 6: SHAP feature importance showing mean absolute SHAP values for brightness (Feature 0) and eye movement variance (Feature 1).*

Figure 7 provides a more detailed view of SHAP values, showing how individual feature values impact model output. Brightness shows a wider distribution of SHAP impacts, confirming its role as the primary driver of risk predictions.

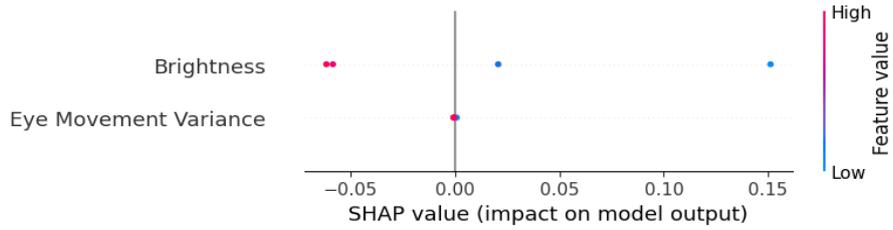

*Figure 7: SHAP value distribution showing the impact of each feature on model output. Color indicates feature value (pink = high, blue = low).*

Table 2: Base Filter Recommendations

| Brightness | Eye Movement Var. | Suggested Filter | Note |
|---|---|---|---|
| High | High | Dark Amber | Maximum protection |
| High | Low | Neutral Density | Minor brightness reduction |
| Medium | High | Cool Grey | Moderate instability |
| Medium | Low | Light Grey | Moderate brightness |
| Low | High/Low | No Filter | Natural vision sufficient |

## 5. DISCUSSION

### 5.1 Limitations

This prototype system has several notable limitations that must be addressed before clinical deployment:

**Synthetic Data Dependency:** The model is trained primarily on synthetic data, which may not capture all nuances of real-world eye movement patterns or environmental brightness transitions. Real patient data would be essential for model validation and refinement.

**Generalization Challenges:** While explainability techniques improve clinical trust, the model's performance on real patients under dynamically changing environments remains unvalidated. Individual variation in photosensitivity thresholds may require personalized model calibration.

**Static Recommendation Engine:** Risk adaptation currently relies on predefined thresholds and fixed filter mappings, lacking the fluidity required for highly personalized, context-sensitive real-world recommendations.

**Device Integration Complexity:** Practical deployment through wearable devices like smart glasses faces hardware constraints including power efficiency, camera sensitivity, and user comfort considerations [17, 18].

### 5.2 Future Directions

To address these limitations and advance toward clinical utility, we propose the following future enhancements:

- **Real Patient Data Collection:** Collaborating with ophthalmology and neurology clinics to gather eye-tracking data and ambient brightness readings from actual patients.
- **Dynamic Feedback Loops:** Incorporating real-time user feedback using reinforcement learning for adaptive behavior [19].
- **Wearable Integration:** Building lightweight, edge-optimized model versions for smart glasses deployment.
- **Context-Aware Adaptation:** Extending recommendations to consider weather, time of day, and indoor/outdoor transitions.

## 6. CONCLUSION

This paper demonstrates the feasibility of an AI-driven personalized assistive system for nystagmus patients facing photosensitivity challenges. We presented a complete end-to-end solution encompassing synthetic data generation, dual-branch deep learning model design, explainability integration via SHAP and GradCAM, and real-time recommendation deployment through a command-line demonstration.

The dual-branch CNN architecture effectively combines environmental brightness and eye movement variance to predict photosensitivity risk scores with 75% validation accuracy. Integration of explainability techniques provides clinical interpretability essential for healthcare AI adoption. The complete implementation is available as open-source software with comprehensive documentation and testing infrastructure.

While significant challenges remain in transitioning from prototype to clinical deployment—particularly around real-world validation and hardware integration—this work establishes a foundational framework for AI-assisted visual risk prediction. Continuous feedback, real-world validation, and iterative technological advancements are essential steps toward transforming this prototype into a life-enhancing tool for the millions of individuals affected by nystagmus and photosensitivity worldwide.

## ACKNOWLEDGMENTS

This work was conducted as part of the AI in Healthcare project at The University of Texas at Austin. The complete implementation is available at: https://github.com/knkarthik01/nystagmus-photosensitivity-ai

## REFERENCES

[1] Y. Lee, S. Lee, J. Han, Y. J. Seo, and S. Yang, "A nystagmus extraction system using artificial intelligence for video-nystagmography," Scientific Reports, vol. 13, no. 1, p. 11975, 2023.

[2] B. Gurnani et al., "Nystagmus in Clinical Practice: From Diagnosis to Treatment—A Comprehensive Review," Clinical Ophthalmology, vol. 19, pp. 1617-1657, 2025.

[3] R. V. Abadi, "Mechanisms underlying nystagmus," Journal of the Royal Society of Medicine, vol. 95, no. 5, pp. 231-234, 2002.

[4] C. Cho, S. Park, S. Ma, H. J. Lee, E. C. Lim, and S. K. Hong, "Feasibility of video-based real-time nystagmus tracking: a lightweight deep learning model approach using ocular object segmentation," Frontiers in Neurology, vol. 15, p. 1342108, 2024.

[5] S. B. Mun, Y. J. Kim, J. H. Lee, G. C. Han, S. H. Cho, S. Jin, and K. G. Kim, "Deep Learning-Based Nystagmus Detection for BPPV Diagnosis," Sensors, vol. 24, no. 11, p. 3417, 2024.

[6] P. Kellnhofer et al., "Gaze360: Physically unconstrained gaze estimation in the wild," in Proc. IEEE/CVF International Conference on Computer Vision (ICCV), 2019, pp. 6912-6921.

[7] S. M. Lundberg and S. I. Lee, "A Unified Approach to Interpreting Model Predictions," in Advances in Neural Information Processing Systems, vol. 30, 2017.

[8] R. R. Selvaraju, M. Cogswell, A. Das, R. Vedantam, D. Parikh, and D. Batra, "Grad-CAM: Visual Explanations from Deep Networks via Gradient-based Localization," in Proc. IEEE International Conference on Computer Vision (ICCV), 2017, pp. 618-626.

[9] H. Zhang and K. Ogasawara, "Grad-CAM-Based Explainable Artificial Intelligence Related to Medical Text Processing," Bioengineering, vol. 10, no. 9, p. 1070, 2023.

[10] K. Noor et al., "Unveiling Explainable AI in Healthcare: Current Trends, Challenges, and Future Directions," WIREs Data Mining and Knowledge Discovery, 2025.

[11] E. Waisberg et al., "Meta smart glasses—large language models and the future for assistive glasses for individuals with vision impairments," Eye, vol. 38, pp. 1036-1038, 2024.

[12] P. Kumar et al., "Artificial intelligence-powered smart vision glasses for the visually impaired," Indian Journal of Ophthalmology, vol. 73, no. 6, pp. 890-897, 2025.

[13] W. Li, Y. Zhang, G. Wang, Y. Huang, and R. Li, "DFENet: A dual-branch feature enhanced network integrating transformers and convolutional feature learning for multimodal medical image fusion," Biomedical Signal Processing and Control, vol. 80, p. 104402, 2023.

[14] W. Wang, J. He, and H. Liu, "EMOST: A dual-branch hybrid network for medical image fusion via efficient model module and sparse transformer," Computers in Biology and Medicine, vol. 179, p. 108771, 2024.

[15] M. T. Ribeiro, S. Singh, and C. Guestrin, "'Why Should I Trust You?': Explaining the Predictions of Any Classifier," in Proc. ACM SIGKDD International Conference on Knowledge Discovery and Data Mining (KDD), 2016, pp. 1135-1144.

[16] S. Bach, A. Binder, G. Montavon, F. Klauschen, K. R. Müller, and W. Samek, "On Pixel-Wise Explanations for Non-Linear Classifier Decisions by Layer-Wise Relevance Propagation," PLOS ONE, vol. 10, no. 7, p. e0130140, 2015.

[17] X. Zhang et al., "Advancements in Smart Wearable Mobility Aids for Visual Impairments: A Bibliometric Narrative Review," Healthcare, vol. 12, no. 24, p. 2594, 2024.

[18] A. Lavric, C. Beguni, and E. Zadobrischi, "A Comprehensive Survey on Emerging Assistive Technologies for Visually Impaired Persons," Sensors, vol. 24, no. 15, p. 4834, 2024.

[19] N. Wagle et al., "aEYE: a deep learning system for video nystagmus detection," Frontiers in Neurology, vol. 13, p. 963968, 2022.

[20] O. Ronneberger, P. Fischer, and T. Brox, "U-Net: Convolutional Networks for Biomedical Image Segmentation," in Medical Image Computing and Computer-Assisted Intervention (MICCAI), 2015, pp. 234-241.